\documentclass[journal]{IEEEtran}
\usepackage[utf8]{inputenc}
\usepackage{array}
\usepackage{amsmath}
\usepackage{amssymb}
\usepackage[T1]{fontenc}
\usepackage{multicol}
\usepackage{textcomp}
\usepackage{filecontents,lipsum}
\usepackage[noadjust]{cite}
\usepackage{cprotect}
\usepackage{fvextra}
\usepackage[dvipsnames]{xcolor}
\usepackage{graphicx}
\usepackage{subfig}
\usepackage[ruled,vlined]{algorithm2e}
\usepackage{multicol,tabularx,capt-of}
\usepackage{multirow}
\usepackage{float}
\usepackage{stackengine}
\usepackage[hidelinks]{hyperref}
\usepackage{hyperref}
\usepackage{accents} 
\usepackage[font=footnotesize]{caption}
\usepackage{colortbl}
\usepackage{scalerel}
\usepackage{tikz}
\usepackage{xcolor,cite,etoolbox}
\usepackage{tikz}
\newcommand{\tikzxmark}{%
\tikz[scale=0.17] {
    \draw[line width=0.58,line cap=round] (0,0) to [bend left=6] (1,1);
    \draw[line width=0.58,line cap=round] (0.2,0.85) to [bend right=3] (0.8,0.05);
}}

\makeatletter 
\pretocmd\@bibitem{\color{black}\csname keycolor#1\endcsname}{}{\fail}
\newcommand\citecolor[1]{\@namedef{keycolor#1}{\color{blue}}}
\makeatletter
\newcommand{\ostar}{\mathbin{\mathpalette\make@circled\star}}
\newcommand{\make@circled}[2]{%
  \ooalign{$\m@th#1\smallbigcirc{#1}$\cr\hidewidth$\m@th#1#2$\hidewidth\cr}%
}
\newcommand{\smallbigcirc}[1]{%
  \vcenter{\hbox{\scalebox{0.77778}{$\m@th#1\bigcirc$}}}%
}
\makeatother
\usetikzlibrary{svg.path}
\definecolor{orcidlogocol}{HTML}{A6CE39}
\tikzset{
  orcidlogo/.pic={
    \fill[orcidlogocol] svg{M256,128c0,70.7-57.3,128-128,128C57.3,256,0,198.7,0,128C0,57.3,57.3,0,128,0C198.7,0,256,57.3,256,128z};
    \fill[white] svg{M86.3,186.2H70.9V79.1h15.4v48.4V186.2z}
                 svg{M108.9,79.1h41.6c39.6,0,57,28.3,57,53.6c0,27.5-21.5,53.6-56.8,53.6h-41.8V79.1z M124.3,172.4h24.5c34.9,0,42.9-26.5,42.9-39.7c0-21.5-13.7-39.7-43.7-39.7h-23.7V172.4z}
                 svg{M88.7,56.8c0,5.5-4.5,10.1-10.1,10.1c-5.6,0-10.1-4.6-10.1-10.1c0-5.6,4.5-10.1,10.1-10.1C84.2,46.7,88.7,51.3,88.7,56.8z};
  }
}
\newcommand\orcidicon[1]{\href{https://orcid.org/#1}{\scalerel*{
\begin{tikzpicture}%[yscale=-1]
\pic{orcidlogo};
\end{tikzpicture}
}{|}}}

\begin{document}

\title{Efficient Object Detection in Optical Remote Sensing Imagery via Attention-based Feature Distillation}

\author{Pourya~Shamsolmoali, \IEEEmembership{Member, IEEE,}
            Jocelyn~Chanussot, \IEEEmembership{Fellow, IEEE,} Huiyu~Zhou, and Yue Lu, \IEEEmembership{Senior Member, IEEE}
             \thanks{

Manuscript received 23 January 2023; accepted 28 October 2023. \\P.~Shamsolmoali and Y.~Lu ({\it Corresponding author}) are with the School of Communication and Electronic Engineering, East China Normal University, Shanghai 200241, China. (Emails: (pshams, ylu)@cee.ecnu.edu.cn).\\
J. Chanussot is with the GIPSA-lab, Universit\'e Grenoble Alpes, CNRS, Grenoble INP 38000, France. (Email: jocelyn.chanussot@grenoble-inp.fr).\\
H. Zhou is with the School of Computing and Mathematical Sciences, University of Leicester, Leicester LE1 7RH, United Kingdom. (Email: hz143@leicester.ac.uk).}}
\markboth{IEEE Transactions on GEOSCIENCE AND REMOTE SENSING}%
{Shamsolmoali \MakeLowercase{\textit{et al.}}}

%and also with the Faculty of Electrical and Computer Engineering, University of Iceland, 101 Reykjavik, Iceland 

% make the title area
\maketitle

\begin{abstract}
Efficient object detection methods have recently received great attention in remote sensing. Although deep convolutional networks often have excellent detection accuracy, their deployment on resource-limited edge devices is difficult. Knowledge distillation (KD) is a strategy for addressing this issue since it makes models lightweight while maintaining accuracy. However, existing KD methods for object detection have encountered two constraints. First, they discard potentially important background information and only distill nearby foreground regions. Second, they only rely on the global context, which limits the student detector's ability to acquire local information from the teacher detector. To address the aforementioned challenges, we propose Attention-based Feature Distillation (AFD), a new KD approach that distills both local and global information from the teacher detector. To enhance local distillation, we introduce a multi-instance attention mechanism that effectively distinguishes between background and foreground elements. This approach prompts the student detector to focus on the pertinent channels and pixels, as identified by the teacher detector. Local distillation lacks global information, thus attention global distillation is proposed to reconstruct the relationship between various pixels and pass it from teacher to student detector. The performance of AFD is evaluated on two public aerial image benchmarks, and the evaluation results demonstrate that AFD in object detection can attain the performance of other state-of-the-art models while being efficient.
%\footnote{The source code will be released at https://github.com/pshams55/REFIPN.}. 
\end{abstract}


\IEEEpeerreviewmaketitle

\section{Introduction}
\IEEEPARstart {R}{ecently}, due to the advancement of deep convolution neural networks (CNNs), significant progress has been made in object detection in remote sensing images \cite{shamsolmoali2021multi, zareapoor2021rotation, liu2021abnet, wu2022uiu}. Nevertheless, most of cutting-edge CNNs, require a large amount of processing power, preventing them from being used on mobile phones and embedded systems. Knowledge Distillation (KD) \cite{hinton2015distilling}, Weight pruning \cite{guo2016dynamic}, and model quantization \cite{jacob2018quantization}, are a few examples of the model compression strategies developed to address this problem.
%%%%%%%%%%%%%%%%%%%
KD in particular has gained popularity as a method for both model compression and model accuracy improvement because of its simplicity and efficacy. In the KD \cite{hinton2015distilling, chen2017learning, zheng2022localization}, a heavyweight teacher network's prediction logits are used to train a smaller, more manageable student network. Therefore, the teacher network's soft labels can assist the student network in making decisions like the teacher network, leading to better performance despite the student network's relatively few parameters.

\begin{figure}
\footnotesize{
  \centering
  \includegraphics[width=0.46\textwidth]{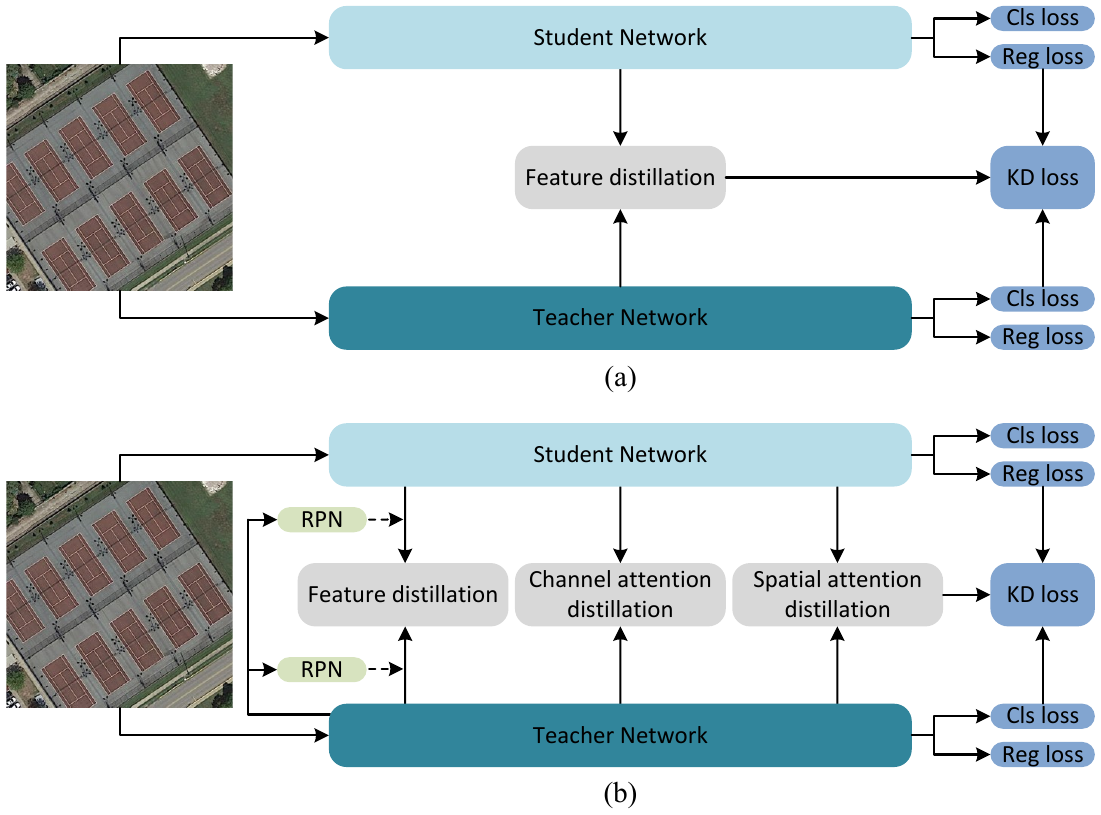}
\caption{KD detection pipelines. (Top) Conventional approaches. (Bottom) Our AFD method. The AFD focuses on obtaining information from the teacher network on both local and global basis.}
\label{fig:1}
}
\end{figure}

The detection of objects and classification of object types in remote sensing images is complicated due to the presence of multiple objects distributed across various locations. This results in vagueness and imbalance in the details of detection. The representations of different positions, such as background, foreground, centers, or borders, may have varying contributions, making the task of KD challenging. The conventional KD approaches \cite{park2019relational, chen2021wasserstein, tung2019similarity} were established for the classification tasks (see Fig. \ref{fig:1}(a)), due to a lack of localization performance, cannot be used for the detection tasks. For example, hint learning \cite{romero2014fitnets} is suggested to distill the transitional feature maps, but it does not pass the localization and classification knowledge of the teacher detector to the student detector. In order to address this concern, \cite{chen2017learning} introduces a new approach to object detection that improves feature extraction, information localization, and classification. Still, because of the disparity between the background and foreground, \cite{chen2017learning} is not able to efficiently extract the teacher's knowledge. In \cite{wang2019distilling}, a feature distillation method is developed, which uses ground truth to filter background regions in order to only perform distillation from the efficient foreground regions. However, this solution does not solve the issue of assigning equal weights to different target regions. Consequently, in \cite{zhang2020improve}, the authors suggest applying mechanism of attention to global features in order to build soft weighted masks, whereby these masks facilitate the access of information from certain and highly important locations. 
However, we have noticed two main problems that arise when relying only on global feature contexts, potentially resulting in the loss of important information within the teacher's features. Firstly, there is a tendency to primarily concentrate on foreground areas while disregarding the background. Neglecting the background is unfavorable for accurate object detection in remote sensing images \cite{zhang2021learning, yang2022adaptive} as it contains valuable information that should not be overlooked. Therefore, efficiently balancing and using all information from both the foreground and the background is the key to boosting distillation performance in object detection. Second, some significant local features that are uniformly distributed in all regions might be overlooked given that the global mask-based approaches just pay attention to the features' global contexts. Applying the softmax function to the global region, would produce an enhanced mask that supplies considerable attention to a foreground object while ignoring the other objects and background areas \cite{zhang2020improve}.

To detect and classify objects in remote sensing images, we propose Attention-based Feature Distillation (AFD) to address the above constraints, as illustrated in Fig. \ref{fig:1}(b). In AFD, we propose a new multi-instance attention strategy that is based on the detector's local and global context features. AFD applies an attention mechanism to local and global features to generate attention masks. In this procedure, the model estimates the attention of various channels and pixels of the teacher's feature map, enabling the student detector to more focus to the teacher's most significant channels and pixels. It also distills the relationship between various pixels from the teacher network and feeds it to the student network. To further extract the teacher's information, the created mask is applied on the extracted features, the Region Proposal Network (RPN) features, the classification output, and the regression output.

Additionally, we incorporate a feature map normalization technique and minimize the MSE loss between the normalized features. This approach aims to mitigate the adverse impact of magnitude disparities between the teacher and student detectors, as well as variations between different Feature Pyramid Network (FPN) layers and channels. In our AFD model, all loss functions exclusively operate on features, allowing for direct integration with different one/two-stage detectors. To evaluate AFD's performance, on two challenging benchmark aerial datasets (DOTA \cite{xia2018dota}, NWPU VHR-10 \cite{dong2019sig}), a comprehensive set of experiments were conducted. The results show that AFD outperforms state-of-the-art KD approaches in object detection. The following is a summary of this paper's significant inventions and contributions:

\begin{itemize}
\item{We introduce an attention-based model for distilling both local and global information from the teacher detector. As a result, student detector focuses more to the foreground objects and less to the background pixels.}
\item{We introduce local and global distillation to enhance the student detector's attention to important teacher channels and pixels, while also fostering an understanding of pixel relationships.}
\item{Comprehensive experiments conducted on two challenging benchmark datasets to thoroughly evaluate our approach. The results demonstrate impressive improvements over other detectors. To illustrate the impact of each module on our propose model's performance, we also performed a comprehensive ablation study.}
\end{itemize}

The rest of this paper is structured as follows. Section \ref{sec:2} dedicated to the brief review of CNN and KD-based object detection methods in natural and remote sensing images. Section \ref{sec:3} describes the proposed AFD model. The dataset details, experimental and evaluation results are given in Section \ref{sec:4}. Section \ref{sec:5} concludes the paper.

\section{RELATED WORK}
\label{sec:2}
Given wide references on object detection models, we focus only on the most recent and closely relevant studies including CNN-based object detection and KD methods.

\subsection{Object Detection}

Current CNN-based object detection models, whether one-stage \cite{liu2016ssd, lin2017focal, zhang2020bridging} or two-stage \cite{he2015spatial, girshick2015fast, ren2016faster}, need considerable processing resources to achieve desired performance, making them impractical for use on embedded devices with limited computation power. These detectors often have a strong backbone, such as VGGs \cite{simonyan2014very} and ResNets \cite{he2016deep}. Consequently, some researches focus on creating lightweight backbone. MobileNet \cite{howard2017mobilenets} is a lightweight deep neural network that using depth-separable convolutions with a complementing search strategie. Single Shot multibox Detector (SSD) \cite{liu2016ssd}, MobileNetV2-SSD \cite{sandler2018mobilenetv2} and MobileNetV3 \cite{howard2019searching} are three examples of lightweight detectors created by combining MobileNet with one-stage detectors.

Existing object detection approaches often rely on adapting image classification frameworks \cite{simonyan2014very, he2016deep} to tackle detection tasks. But since classification and detection tasks are so distinct from one another, a lightweight backbone is not ideal for direct deployment. Hence, some lightweight detectors like Tiny-deeply supervised object detection \cite{li2018tiny} and Pelee \cite{wang2018pelee} have developed specific backbones. To accomplish effective real-time detection, ThunderNet \cite{qin2019thundernet} proposes integrating a compacted backbone with a RPN. 
%%%%
\begin{figure*}
  \centering
  \includegraphics[width=0.83\textwidth]{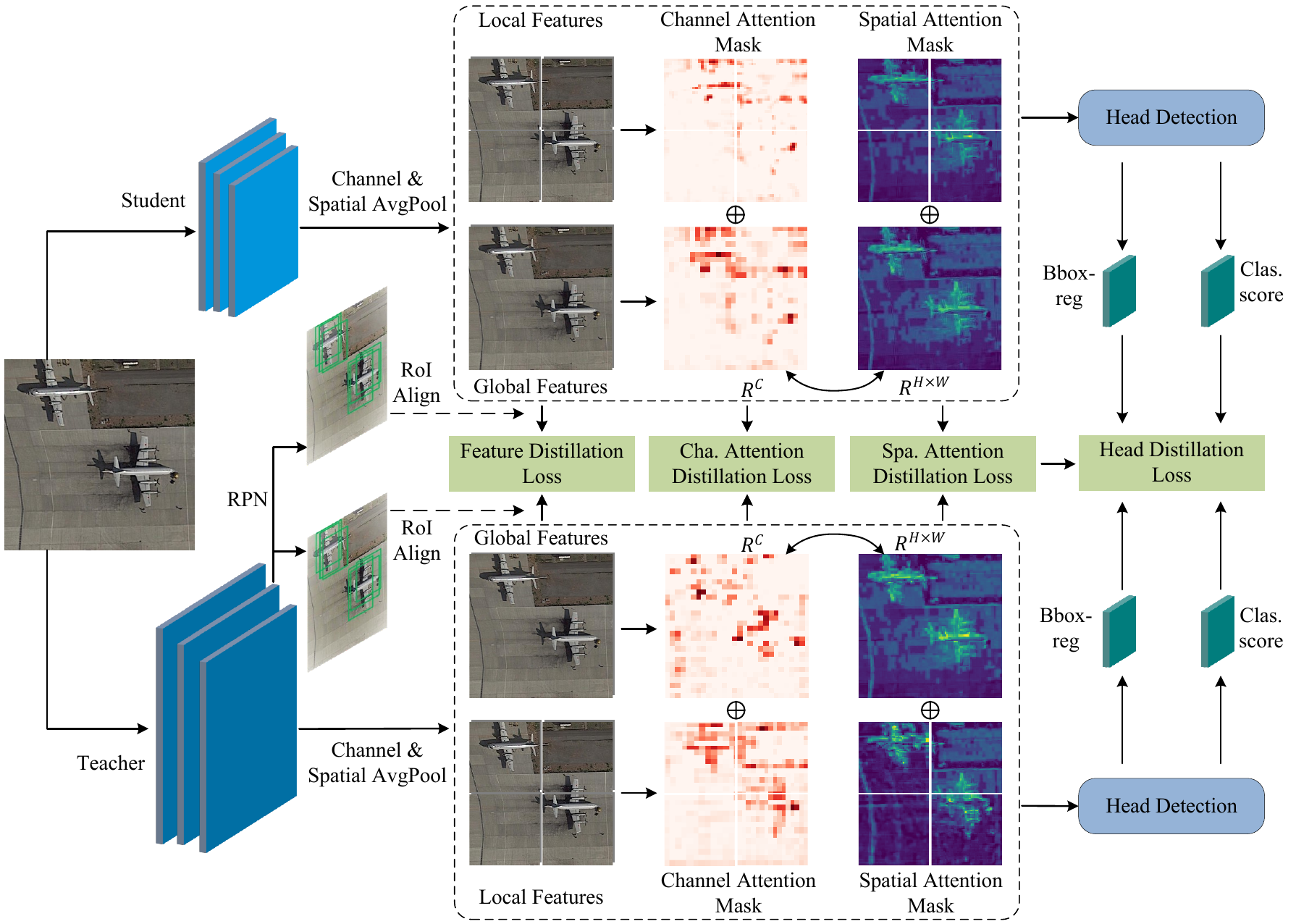}
\caption{Architecture of our KD method. The enhancement of AFD is based on three points. (a) Our new KD approach distils both local and global information from the teacher network. (b) For local distillation, a multi-instance attention mechanism is proposed to identify the background from the foreground. (c) Attention local and global distillation is proposed to reconstruct the relationship between various pixels and pass it from teacher to student detector.}
\label{fig:2}
\end{figure*}
%%%%%%%

These lightweight detectors often do not provide good detection results when used for remote sensing imagery due to its complex background and multiscale objects. In light of this need, several deep learning-based detectors have been proposed for remote sensing images. To better focus on tiny objects, the authors of \cite{wang2019fmssd} propose adopting the atrous spatial feature pyramid component and integrating multiscale context information through a loss weighted by region. Merging attention with deformable convolution for object detection is proposed in \cite{sun2020sraf}, via context-based deformable module on the basis of contextual information. Spatial misalignment between anchors and ground truth is one of the problems of object detection in remote sensing images. To address this issue, in \cite{xu2021assd} a unique pseudo-anchor proposal module is introduced. To efficiently address the problem of rotated objects, \cite{zareapoor2021rotation} proposes a method that learns rotation invariants and trains the network by applying additional constraints. In addition, to boost geospatial recognition accuracy, \cite{li2017rotation} proposes a network based on local-contextual feature fusion and \cite{shamsolmoali2021multi} introduce a new model to enhance feature maps quality for better object detection. In \cite{wu2021deep}, a survey of object detection and tracking methods for remote sensing images is gathered that provides thoughts for models further development.
In \cite{shamsolmoali2022enhanced}, a pyramid single-shot detector is proposed for small object detection in remote sensing images. To further enhance small object detection, in \cite{wu2022infrared}, an interactive U-Net architecture is proposed which has higher feature learning by utilising object's global context information. In \cite{wu2019orsim}, a detector is proposed that uses spatial-frequency channel features by incorporating both rotation-invariant channel features and original spatial channel features which enhances the system's robustness, and accuracy. However, these detectors are difficult to deploy on devices with limited storage and computational power.

\subsection{KD for Object Detection}
To develop precise and lightweight detectors for natural scenes, researchers have extensively utilize KD in recent years \cite{chen2017learning}. The application of KD in this particular task focuses on distilling the distinct locations of the detector. However, during the process of imitating the feature maps, the imbalanced distribution of foreground and background pixels is often disregarded, leading to inferior performance. To solve this problem, the authors of \cite{wang2019distilling} suggest an imitation technique for fine-grained features, which focuses the detector's attention on the objects. To identify the central foreground pixels, a 2-dimension Gaussian mask is applied in the ground-truth regions for feature distillation in \cite{sun2020distilling}. This strategy decreases the imbalance at the expense of eliminating the backgrounds. 
On the other hand, recent analysis \cite{guo2021distilling, dai2021general} has shown that the background areas contain important information. Specially, remote sensing objects are often connected to their environments. During the distillation process, it is important to pay attention to the areas around the objects and the background. In \cite{yang2022focal}, Focal and Global Distillation is proposed that consists of, focal distillation for foreground-background separation, and global distillation for pixel relationship restoration.
\vspace{-1pt}

In \cite{chen2017learning}, the authors integrate the boundary regression loss of teacher detector with the regression elements and the transmission of unbounded regression data, which lacks distinctions between objects with varied levels of difficulty in regression. In \cite{sun2020distilling}, a successful classification and regression model is designed using the $\ell_1$ and binary cross-entropy losses. These distillation approaches continue to underestimate the importance of background, which result in loss of contextual information around the objects. \cite{dai2021general} introduces a distillation mask-based method that focuses on discriminative patches by determining the differences between the teacher and student results. Moreover, in order to restrict the feature maps, a multiscale feature transition on the output of the FPN is applied. Similarly, in \cite{zhang2020improve}, a soft mask-based method is developed which extracts feature attention from its backbone. Conversely, current global attention masks generally overlook other significant regions as networks primarily have attention to small objects or regions. In order to further boost detection performance, we propose creating attention masks for all the local patches. These masks would direct attention to other important patches that contain local information.

\section{METHODOLOGY}
\label{sec:3}
In this section, we outline the details of our local and global attention mask to accurately represent the characteristics of features. Then describe how the feature distillation and head distillation are accomplished. Fig. \ref{fig:2} represents the overview of our proposed AFD method.

\subsection{Local and Global Attention-based Mask}
A fundamental part of the proposed AFD is the local and global attention-based masks (LGAM), which we discuss below. LGAM incorporates channel attention $M_{cha}$ with spatial attention $M_{spa}$ methods. To get the channel attention masks, a softmax is applied to the channel dimension, as the average weight of the feature components $|x_{i,j}|$ over the channel dimension. The proposals $\mathcal{P}_x^T$ generated by the RPN of teacher are shared with the teacher and the student detector in order to obtain the same candidates for loss computation between the detectors. The RPN module has a positive impact on proposal quality, localization accuracy, and efficiency. It allows the student detector to benefit from the teacher's knowledge, leading to improved object detection performance.
%%%%%%%%%%%%%
\begin{equation}
 \begin{aligned}
M_{cha}(x)=HW.\rho~(\frac{\frac{1}{HW}.\sum_{i=1}^H \sum_{j=1}^W(|x_{i,j}, \mathcal{P}_x^T|)}{\mathcal {T}}),
\label{eq:1}
 \end{aligned}
\end{equation}
%%%%%%%%%%%%
in which $\rho(.)$ is the softmax operation and $\mathcal T$ denotes the temperature parameter. For an input feature, $H$ and $W$ denote its height and width. Consequently, the channel-wise feature components $|x_k|$ is utilized in the operations of softmax with the $H$ and $W$ dimensions to generate the spatial attention masks as written below:
%%%%%%%%%%%%%
\begin{equation}
 \begin{aligned}
M_{spa}(x)=C.\rho~(\frac{\frac{1}{C}.\sum_{c=1}^C(|x_{c}, \mathcal{P}_x^T|)}{\mathcal {T}}),
\label{eq:2}
 \end{aligned}
\end{equation}
%%%%%%%%%%%%
in which $C$ denotes the feature's channel of input. To create LGAM that incorporate both local and global perspectives, we divide each FPN output feature into $P$ local features $f_p\in \mathbb {R}^{I\times I\times C}$, in which $I$ is the predetermined instance size and $p \in\{1, 2, ..., P\}$. Consequently, we can write the local channel and spatial masks ($L_{ch}$, $L_{sp}$) as:
%%%%%%%%%
\begin{equation}
\begin{aligned}
L_{ch},p = M_{cha}(f_p^T)+ M_{cha}(f_p^S),~L_{ch}=\otimes(L_{ch,1},L_{ch,2},\\
..., L_{ch,P},)
\label{eq:3}
\end{aligned}
\end{equation}
%%%%%%%%%%
\begin{equation}
\begin{aligned}
L_{sp},p = M_{spa}(f_p^T)+ M_{spa}(f_p^S),~L_{sp}=\otimes(L_{sp,1},L_{sp,2},\\
..., L_{sp,P},)
\label{eq:4}
\end{aligned}
\end{equation}
%%%%%%%%%%%
%%%%%%%%%%%%
\begin{figure}
  \centering
  \includegraphics[width=0.44\textwidth]{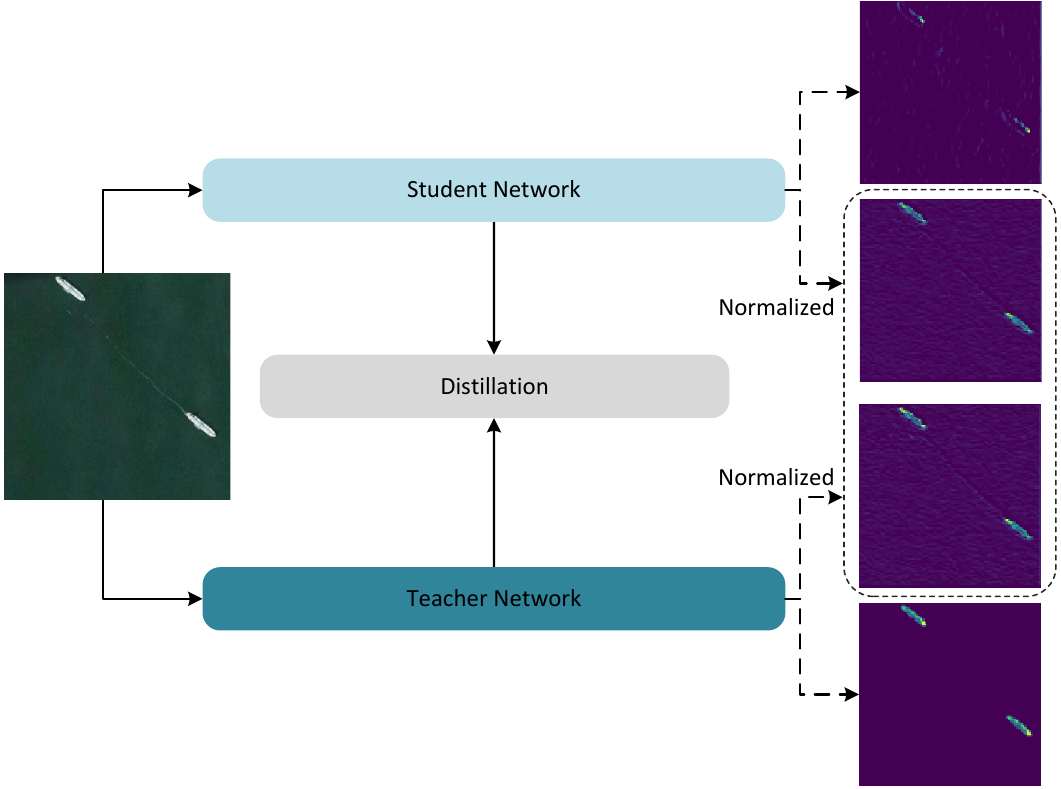}
\caption{Activations for the input image before and after normalization. This process fills the gap between the patterns of the teacher detector and the student detector, providing a more effective and smoother transfer of knowledge.}
\label{fig:3}
\end{figure}
%%%%%%%%%%%
in which $T$ stands for teacher, $S$ for student, and $\otimes$ for the concatenation operator. An effective feature distillation approach must considers magnitude difference while generating pairs for imitation. In addition, by analyzing the activation patterns, we observe that the dominating FPN layers and channels may directly interact with the student's training phase and lead to sub-optimal performance, which is ignored by previous studies. To overcome this problem and optimize the learning process as shown in Fig. \ref{fig:3}, we suggest first normalize the teacher's features and the student ones. This involves transforming the features to have a zero mean and a unit variance. Once the normalization is complete, the next step is to minimize the mean squared error (MSE) between the normalized features. It is also important that the normalization follow the convolution property, to ensure that features are normalized uniformly at different regions of the feature map.
Let $\mathbb V$ represent the whole set of feature map values that include the components of mini-batch and its spatial locations. Therefore, for an $u$-size mini-batch and $h\times w$-size feature maps we take the functional mini-batch of $m = ||\mathbb V|| =u-size.hw$. Let $s^{(c)}\in \mathbb R^m$ be the $c^{th}$ channel in a batch of FPN outputs, therefore, we can obtain the normalized values from the teacher $T$ and the student $S$ detector. Consequently, in the same way that local features are normalized, the global feature $\mathcal {F} \in \mathbb R^{H \times W\times C}$ can be normalized. Therefore, the global channel and spatial masks ($G_ch$, $G_sp$) can be written as:
%%%%%%%%%%
\begin{equation}
\begin{aligned}
G_{ch} = M_{cha}(\mathcal {F}^T)+ M_{cha}(\mathcal {F}^S),\\
G_{sp} = M_{spa}(\mathcal {F}^T)+ M_{spa}(\mathcal {F}^S).
\label{eq:5}
\end{aligned}
\end{equation}
%%%%%%%%%%%
In order to build our final channel attention masks $LG_{ch}$ and spatial attention masks $LG_{sp}$, we integrate the local and global masks as illustrated below:
%%%%%%%%%%
\begin{equation}
\begin{aligned}
LG_{ch} = \frac{1}{2}.(L_{ch}+ G_{ch}), ~LG_{sp} = \frac{1}{2}.(L_{sp}+ G_{sp}).
\label{eq:6}
\end{aligned}
\end{equation}
%%%%%%%%%%%

%%%%%%%%%%%%
\begin{figure}
  \centering
  \includegraphics[width=0.47\textwidth]{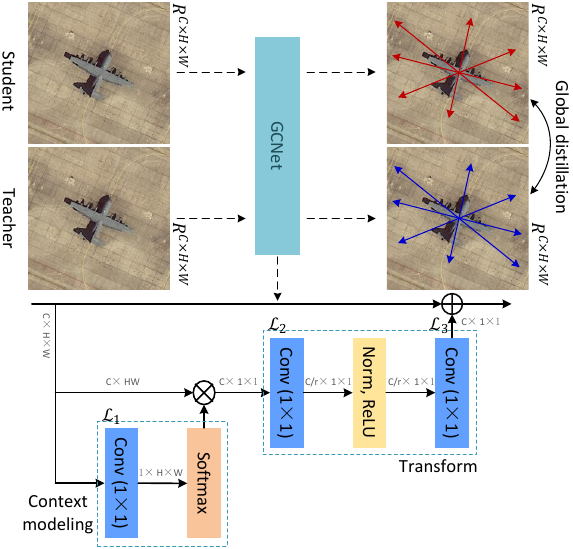}
\caption{The framework of global distillation. The feature maps from the student and teacher detectors used as the inputs.}
\label{global}
\end{figure}
%%%%%%%%%%%

In our model, the feature maps normalization of both student and teacher detectors helps to align the magnitudes, improve knowledge transfer, and enhance stability. These effects collectively contribute to more effective and efficient distillation and result in improved student detector performance.
%---------------------------------
\subsection{Feature-based Distillation}
In KD, the features of teacher detector generally contain more information than the features of student detector. Hence, we distill the FPN's intermediate features to boost students' performance. To carefully distill the region of interest, all layers' features are combined with the spatial and channel attention masks. We define the loss of feature distillation as:
%%%%%%%%%%
\begin{equation}
\begin{aligned}
\ell_{fd} =\sum_{l=1}^L(\sum_{c=1}^C\sum_{i=1}^H\sum_{j=1}^W(\mathcal {F}_{lcij}^T-\phi_{adj}(\mathcal {F}_{lcij}^S))^2.LG_{sp,l}.LG_{ch,l})^{\frac{1}{2}},
\label{eq:7}
\end{aligned}
\end{equation}
%%%%%%%%%%%
in which depth of the FPN is shown by $L$, $l$ stands for the $l^{th}$ FPN layer, while $i$ and $j$ are the locations of the feature map with the corresponding $H$ and $W$. $\phi_{adj}$ is the $1\times1$ convolution layer used to adjust the student's features to those of the teacher. In addition, $LG_{ch,l}$ and $LG_{sp,l}$ are the mean channel and the spatial masks of the $l^{th}$ layer, respectively. Further, the attention features are distilled in order to support the student to generate better LGAM. So, we can write the procedure of channel attention feature and spatial attention feature extraction as $AF_{ch}(x)=\frac{1}{C}.\sum_{c=1}^{C}x_{c}$ and $AF_{sp}(x)=\frac{1}{HW}.\sum_{i=1}^{H}\sum_{j=1}^{W}x_{ij}$. Through the distillation of local and global channel attention features, the channel attention loss is computed. Furthermore, the features obtained from local and global spatial attention are equivalent, while the local features are derived through splitting the global features in the spatial domain. Spatial attention loss, unlike channel attention loss, mainly uses global spatial attention features. Thus, we can write our channel attention and spatial attention losses ($\ell_{cha}$, $\ell_{spa}$) as:
%%%%%%%%%%
\begin{equation}
\begin{aligned}
\ell_{cha} =\frac{1}{2}.({\parallel AF_{ch}(\mathcal {F}^S)-AF_{ch}(\mathcal {F}^T) \parallel}_2 + \\
\frac{1}{N}.\sum_{p=1}^{P}{\parallel AF_{ch}(f_p^S)-AF_{ch}(f_p^T) \parallel}_2),
\label{eq:8}
\end{aligned}
\end{equation}
%%%%%%%%%%%
\begin{equation}
\begin{aligned}
\ell_{spa} ={\parallel AF_{sp}(\mathcal {F}^S)-AF_{sp}(\mathcal {F}^T) \parallel}_2.
\label{eq:9}
\end{aligned}
\end{equation}
%%%%%%%%%%%
Now we can write the feature attention loss $\ell_{fa}$ by combining the channel attention loss and spatial attention loss
%%%%%%%%%%%
\begin{equation}
\begin{aligned}
\ell_{fa} =\ell_{cha} + \ell_{spa}.
\label{eq:10}
\end{aligned}
\end{equation}

%%%%%%%%%%%
\subsection{Global Distillation}

The relationship \cite{cao2019gcnet, wang2018non} between different pixels contains useful information that is used to boost detection task performance. In addition to feature attention, which aims to sever the relationship between background and foreground, a global distillation approach is also proposed. This approach facilitates the transfer of key knowledge from the teacher detector to the student detector by leveraging the global relationships between neighboring pixels in the feature maps.
To compel the student detector to acquire knowledge about the pixels' relationship from the teacher detector, we use GcNet \cite{cao2019gcnet} for the purpose of extracting the global relation information from an image, as demonstrated in Fig. \ref{global}. Consequently, we can write the global loss as:
%%%%%%%%%%
\begin{equation}
\begin{aligned}
\ell_{glob} =\Lambda. \sum(B(\mathcal {F}^T)-B(\mathcal {F}^S))^2,
\label{eq:10-1}
\end{aligned}
\end{equation}
%%%%%%%%%%%
in which $B(\mathcal {F})=\mathcal {F}+\mathcal{L}_3(\mathcal{N}(ReLU(\mathcal{L}_2(\sum_{j=1}^{n^p}\frac{e^{\mathcal{L}_1\mathcal {F}_j}}{e^{\mathcal{L}_1\mathcal {F}_M}}\mathcal {F}_j))))$, $\mathcal{L}_1$, $\mathcal{L}_2$, and $\mathcal{L}_3$ are layers of CNN. $\mathcal{N}$ is the normalizing layer, $n^p$ denotes the number of pixels and $\Lambda$ represents the loss-balancing hyper-parameter.
This hyper-parameter controls the trade-off between performance gains and knowledge transfer in the student detector. It allows fine-tuning the amount of knowledge transferred from the teacher detector while ensuring that the student detector learns effectively from its own training data.
%%%%%%%%%%%%

%%%%%%%%%%%
\begin{table}
\renewcommand{\arraystretch}{1.1} 
\centering
\caption{OUR AFD'S ABILITY FOR GENERALIZATION IN DIFFERENT OBJECT DETECTION MODELS ON THE DOTA AND NWPU. WE COMPARE OUR MODEL'S PERFORMANCE BY mAP (\%), FPS (f/s), AND NUMBER OF PARAMETERS (M).} \label{tab.1}
\begin{tabular}{l|ccc|ccc}     \hline\hline
%\multicolumn{5}{|c|}{Results on training on untransformed data (0)} \\  \hline
\multirow{ 2}{*}{Method} & \multicolumn{3}{c|}{DOTA}& \multicolumn{3}{c}{NWPU} \\ [-0.3ex]  \cline{2-7}   
            & mAP & FPS& Params& mAP & FPS& Params \\ [-0.3ex]   \hline

FR-CNN (T)  &   72.18& 18&92.65 &90.91 & 22& 73.62 \\  [-0.5ex]
FR-CNN (S)  &   65.27& 32&60.17 &85.93 & 39&41.15  \\ [-0.5ex]
AFD   &    70.54 & 32&60.17  &89.76 & 38&41.15 \\   \hline
%------------------------------------------
Cascade (T)  &   77.39& 16&120.24 &93.14 & 19&101.22  \\  [-0.5ex]
Cascade (S)  &   70.47& 30&87.80 &88.42 & 36&68.77  \\ [-0.5ex]
AFD  &  76.91& 30&87.80  &91.85 & 35 &68.77 \\   \hline
%------------------------------------------
RetinaNet (T) &  72.94& 19&87.74  & 90.86& 23& 68.87 \\  [-0.5ex] 
RetinaNet (S) &    64.47& 32&55.36 &85.91 & 41&36.28   \\ [-0.5ex]
AFD  & 73.08&  32& 55.36 &90.93 & 40&36.28 \\   \hline
%------------------------------------------
ATSS (T)  &  74.18& 18&55.45  & 92.90& 22&51.44  \\  [-0.5ex] 
ATSS (S) &    67.42& 33&18.97 &86.52 & 39& 18.95  \\ [-0.5ex]
AFD  & 72.94&  33&18.97  &92.67 & 39&18.95 \\   \hline
%------------------------------------------
FCOS (T)  &  72.56& 19& 67.98 & 91.84& 22& 64.33 \\  [-0.5ex] 
FCOS (S)  &    67.71& 33&31.56 &87.21 & 41& 31.85  \\ [-0.5ex]
AFD  & 71.82&  33&31.56  &90.63 & 40& 31.85\\   \hline
\end{tabular}
\end{table}
%%%%%%%%%%%%%%%%%%%%%%%%%%%%%%%%%

\subsection{Head Distillation}

By directing attention towards the outputs of the students, the distillation process stimulates them to attain performance levels comparable to that of the teacher. Nonetheless, in the case of remote sensing images, where there exists a substantial imbalance between background and foreground, directly distilling the outputs from the teacher's head may adversely affect the detection performance of the student. That's why spatial attention masks are used to ensure that the response-based distillation is as accurate as possible. In particular, from the FPN we take the spatial attention masks (see Eq. (\ref{eq:6})) to preform masked head distillation and we can write the classification head loss $\ell_{cls-h}$ as:
%%%%%%%%%%
\begin{equation}
\begin{aligned}
\ell_{cls-h} =\sum_{l=1}^L\sum_{c=1}^C\sum_{i=1}^H\sum_{j=1}^W\ell_{ce}(o_{lcij}^S,o_{lcij}^T).LG_{sp,l},
\label{eq:11}
\end{aligned}
\end{equation}
%%%%%%%%%%%
in which $o^S$ and $o^T$ denote the classification head's outputs for both student and teacher detectors, and $\ell_{ce}$ is the cross-entropy loss. As stated in \cite{chen2017learning}, the student model receives inappropriate information from unbounded teacher outputs. To address this problem, IoU loss is adopted to distill the localization head and we can define it's loss as:
%%%%%%%%%%
\begin{equation}
\begin{aligned}
\ell_{loc-h} =\sum_{l=1}^L\sum_{c=1}^C\sum_{i=1}^H\sum_{j=1}^W\ell_{IoU}(k_{lcij}^S,k_{lcij}^T).LG_{sp,l},
\label{eq:12}
\end{aligned}
\end{equation}
%%%%%%%%%%%
in which $k$ denotes the output of the localization head.

%\subsection{Overall Loss}
By incorporating the outputs generated by the modules of the detector with the Faster R-CNN \cite{ren2016faster}, we properly set the distillation losses and then compute the total loss by aggregating the $\ell_{cls-h}$ and $\ell_{loc-h}$ for object detection as,
%%%%%%%%%%
\begin{equation}
\begin{aligned}
\ell_{total} =\nu \ell_{fd}+ \upsilon\ell_{fa}+ \ell_{glob}+ \beta(\ell_{cls-h}+\ell_{loc-h})+ \ell_{rpn},
\label{eq:13}
\end{aligned}
\end{equation}
%%%%%%%%%%%
%-----------------------
\begin{figure}
  \centering
  \includegraphics[width=0.47\textwidth]{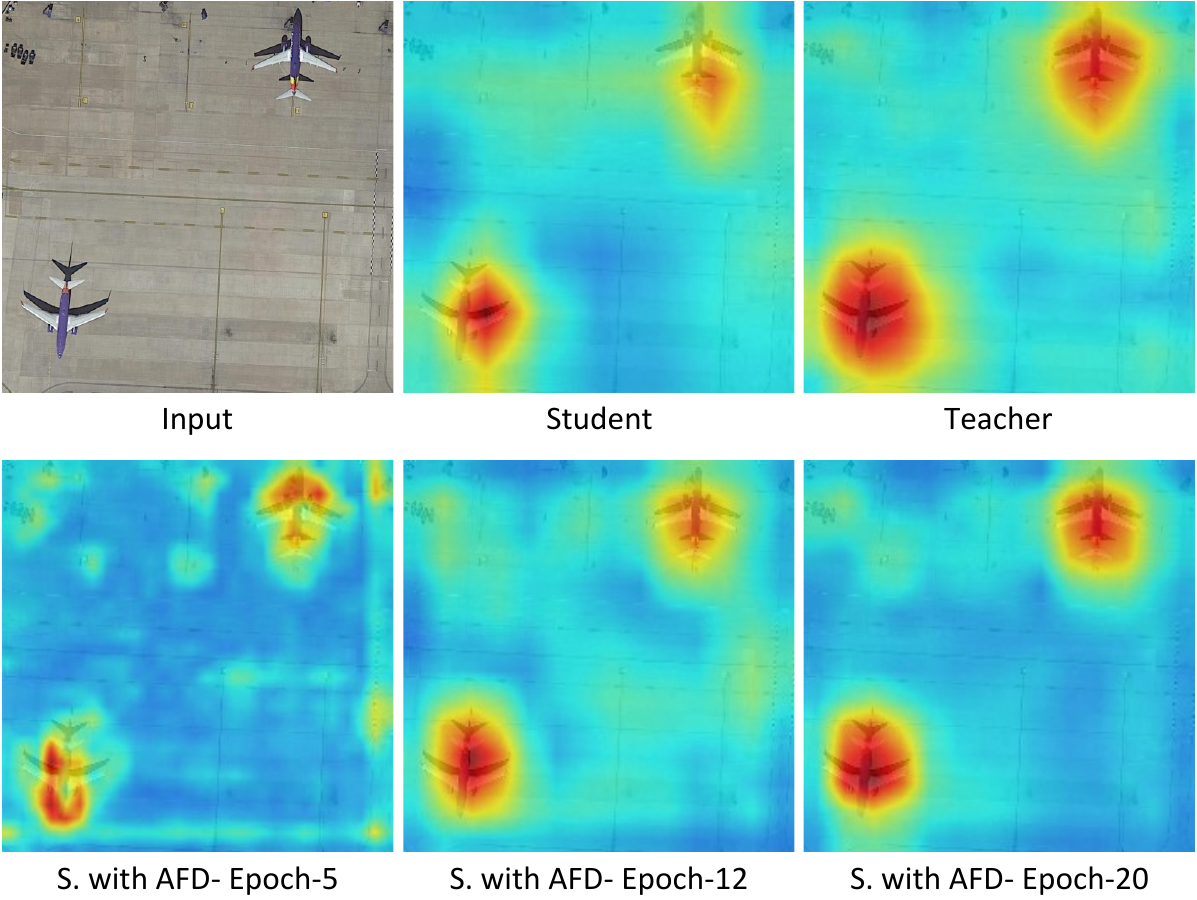}
\caption{{Visualization of the attention maps produced by the student detector, teacher detector, and different training phases of student detector using AFD. Red denotes the highest level of attention, whereas other colors denote lower.}}
\label{attention-2}
\end{figure}
%%%%%%%

\noindent in which $\nu$, $\upsilon$ and $\beta$ represent the balancing-parameters for the different losses. The $\ell_{rpn}$ denotes the loss of RPN \cite{ren2016faster} in two-stage detector which written as:
\begin{equation}
 \begin{aligned}
\ell_{rpn}&=\lambda_1 \frac{1}{\mathcal N_{cls-h}}\sum_i \ell_{cls-h}(p_i, p^{\ast}_i) \\ 
&  +\lambda_2 \frac{1}{\mathcal N_{reg}}\sum_i p^{\ast}_i \ell_{reg}(t_i, t^{\ast}_i)
 \end{aligned}
\end{equation}

\noindent in which $i$ is the index of a bounding box (BB), $p_i$ is the probability of the $i^{th}$ anchor predicted as an object, $p^{\ast}_i$ is the ground-truth type appointed to the $i^{th}$ anchor (0 if the box is negative and 1 for the positive one), $\ell_{reg}$ is the smooth-$\ell_1$ loss, $t_i$ represents the detected regression offset for $i^{th}$ anchor and $t^{\ast}_i$ denotes the target BB regression offset for the $i^{th}$ positive anchor. The hyper-parameters $\lambda_1$ and $\lambda_2$ denote the balancing factors for losses, which we adjusted to 1 in our experiments for simplicity. $\mathcal N_{cls-h}, \mathcal N_{reg}$ denote normalization parameters that help to reduce the effect of various object scales, resulting in more effective training.

\section{EXPERIMENTAL RESULTS AND ANALYSIS}
\label{sec:4}
In this section, we encompass a detailed description of the datasets used, the evaluation metrics employed, and the experiments undertaken to assess the effectiveness and efficiency of our KD approach. Furthermore, to determine the effect of each module on the overall performance of the proposed architecture, a comprehensive ablation study is conducted.

\subsection{Datasets and Evaluation Metrics}
{\it DOTA} \cite{xia2018dota} is a remote sensing image dataset for object detection that contains $2806$ images of various sizes. It consists of $15$ types of objects with various dimensions and orientations.

{\it NWPU VHR-10} \cite{dong2019sig} is a dataset that contains $650$ remote sensing images of different sizes. It consists of $10$ types of objects. %We randomly selected $80\%$ of the original images for training, $10\%$ for validation, and the rest for testing.

For the DOTA dataset, the images are cropped into the $800 \times 800$ pixels patches with 200 pixels overlap with the neighboring patches. 
%In addition, the multi-scale technique is used in this process \cite{xu2020hierarchical}. To be more specific, we first rescale the actual images by $1.5\times$, and $0.5\times$ before splitting, and then use all of the patches for the training and testing.
%-----------------------
\begin{figure}
  \centering
  \includegraphics[width=0.415\textwidth]{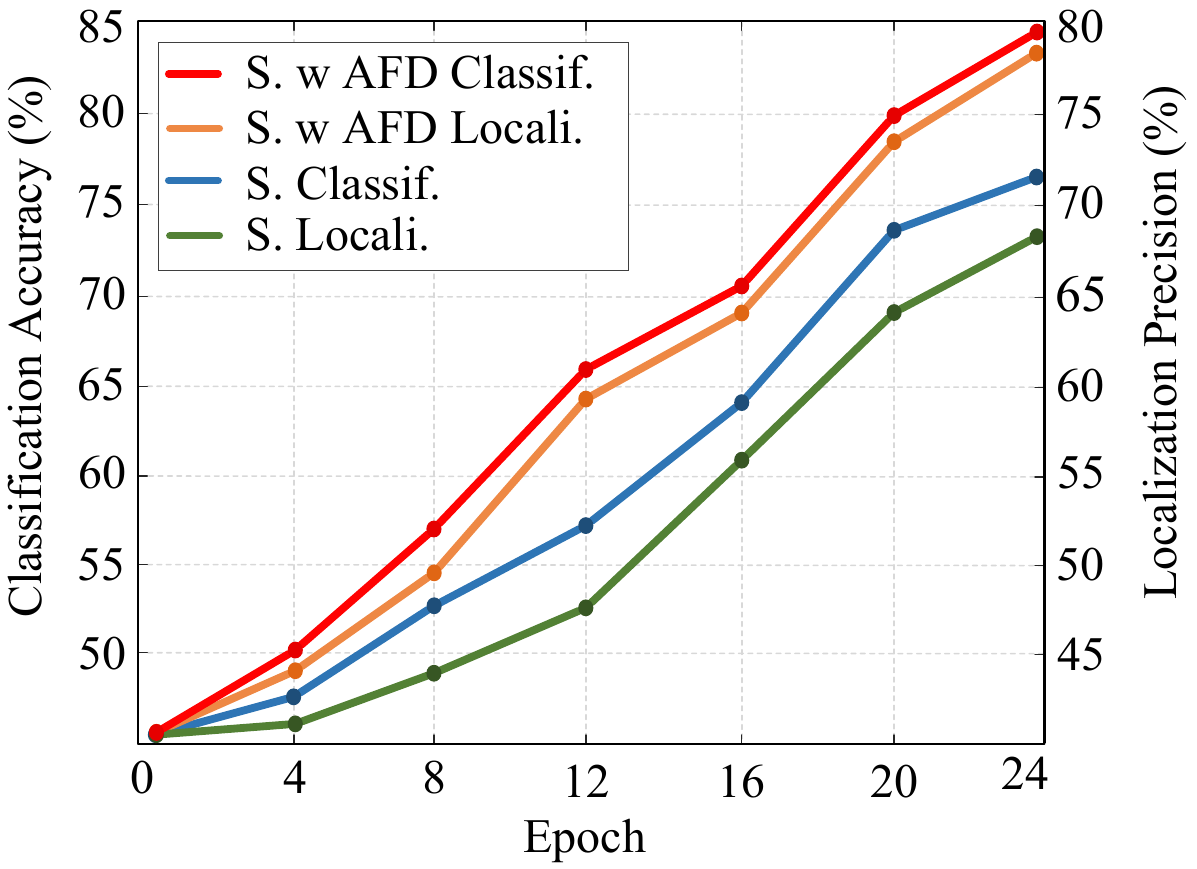}
\caption{{Classification accuracy and localization precision of the student detector on FR-CNN with and without our AFD module on the DOTA dataset during training. }}
\label{Fig7}
\end{figure}
%%%%%%%
For the NWPU dataset there are not enough images for training. For expanding the training dataset, we performed rescaling, rotation and flipping. %Moreover, as the number of the entities in each class was unbalanced, we use different techniques for each class to balance the number of objects. 
%%%%------------

The metrics of {\it mean average precision} (mAP), {\it frames per second} (FPS), and {\it number of parameters} (Params) are used to assess the performance of AFD. Following is how mAP is calculated:
%--------------------------------
\begin{eqnarray}
mAP=\int_{0}^{1} P(R)\; dR,
\label{eq:23}
\end{eqnarray}
%------------------------------
where the predicted rates for accuracy and recall are $P$ and $R$, respectively, and $d$ denotes the coordinates of the estimated center point. 
In addition, to more accurately assess a method's ability for localisation and classification, the metrics of {\it Localization Error} and {\it Confusions with Background} \cite{fu2020point, fu2020rotation} are used.

%%%%%%%
\begin{figure*}
  \centering
  \includegraphics[width=0.94\textwidth]{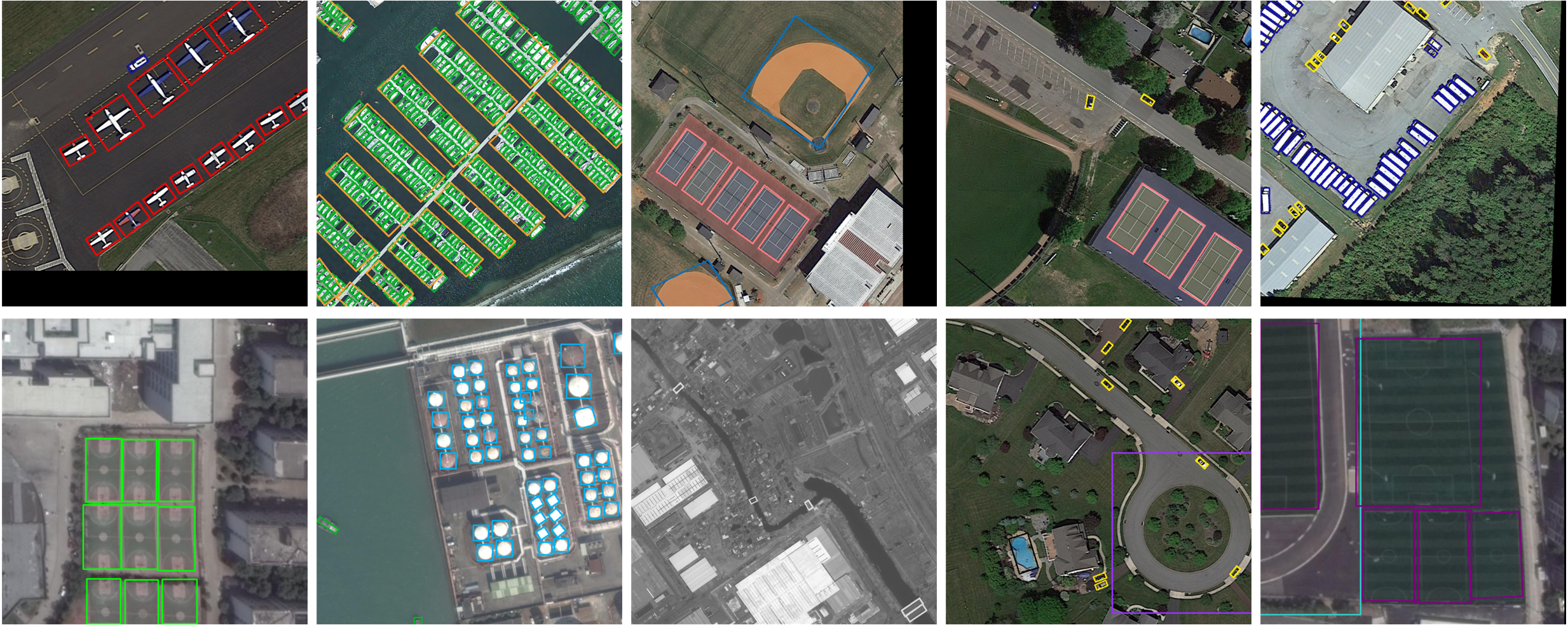}
\caption{Sample detection results of the student using Cascade detector trained with AFD on the DOTA.}
\label{dota}
\end{figure*}
%%%%%%%%
%%%%%%%
\begin{table*}
\centering
  \caption{COMPARISON WITH STATE-OF-THE-ART OBJECT DETECTION KD METHODS ON THE DOTA USING RETINANET.}
  \label{tab.2}
  \begin{tabular}{m{6.3em} |  m{0.5cm} m{0.5cm} m{0.5cm} m{0.5cm} m{0.5cm} m{0.5cm} m{0.5cm} m{0.5cm} m{0.5cm} m{0.5cm} m{0.5cm} m{0.5cm} m{0.6cm} m{0.5cm} m{0.5cm}| m{0.5cm} }     \hline\hline
    Methods &Plane&BD&Bridge&GFT&SV&LV&Ship&TC&BC&ST&SBF&RA&Harbor&SP&HC&mAP  \\     \hline

{Teacher}  &88.96&83.18&52.68&62.67&72.51&73.84&79.42&90.57&81.98&79.82&54.77&67.82&75.49&71.84&58.51& 72.94	\\

{Student} 	  &86.67&72.63&40.58&51.69&71.82&65.61&77.46&89.64&68.22&72.78&41.59&56.42&64.87&69.05&37.98&64.47	 \\\hline
%------------
{FGFI \cite{wang2019distilling}}  &88.45&75.96&44.51&56.30&72.89&63.58&74.96&90.78&76.81&72.19&48.43&61.63&70.06&69.11&45.51&	67.41	 \\ 

{TAR \cite{sun2020distilling}}   &89.16&76.52&46.55&59.98&73.90&66.24&78.56&{{90.78}}&78.57&75.67&43.71&66.93&71.40&{{72.16}}&45.32&69.03	\\ 

%---------------
{DKD \cite{zhang2021learning}}  &89.48&81.48&{{46.38}}&60.52&76.25&64.18&78.36&90.79&78.60&78.31&{{53.12}}&65.06&73.05&\bf{74.11}&{{59.98}}&71.31  \\ 

%-------------------
{FGD \cite{yang2022focal}}  &89.60&81.55&{\bf{47.63}}&{60.34}&76.19&64.26&78.26&{90.46}&78.43&78.45&{{52.81}}&65.20&73.32&73.67&{{60.14}}& 71.36 \\ 
%%----------------------
{LD \cite{zheng2022localization}}  &89.64&81.54&{{46.57}}&\bf{60.73}&76.42&64.15&78.51&\bf{90.84}&78.72&78.41&{{53.26}}&65.23&73.18&73.91&{{60.21}}& 71.43 \\ 
%-------------------

{AFD (ours)} 	 &{\bf{89.91}}&{\bf{82.63}}&{47.59}& 60.58&{\bf{77.30}}&{\bf{65.37}}&\bf{79.14}&90.75&{\bf{79.22}}& \bf{78.84}&\bf{54.56}&{\bf{65.94}}&{\bf{74.49}}&{74.06}&\bf{62.48}& 	{\bf{73.08}}	 \\  \hline
\end{tabular}
\end{table*}
%----------------------------------
%%%%%%%%%%%%%%%%%%%%%%%%%%%%%%%%%%%%
\begin{table*}
\centering
  \caption{COMPARISONS OF DETECTION RATE AND SPEED FOR OUR MODEL WITH CASCADE AGAINST OTHER DETECTORS ON THE DOTA DATASET FOR HBB TASK. THE BEST RESULTS ARE HIGHLIGHTED.}
  \label{tab:3}
  \begin{tabular}{m{6.8em} |  m{0.5cm} m{0.5cm} m{0.5cm} m{0.5cm} m{0.5cm} m{0.5cm} m{0.5cm} m{0.5cm} m{0.5cm} m{0.5cm} m{0.5cm} m{0.5cm} m{0.6cm} m{0.5cm} m{0.5cm}| m{0.5cm} m{0.24cm}}     \hline\hline
    Methods &Plane&BD&Bridge&GFT&SV&LV&Ship&TC&BC&ST&SBF&RA&Harbor&SP&HC&mAP&FPS  \\     \hline
%\multicolumn{18}{c}{\it Detectors for HBB task}  \\ 
%
%{SSD \cite{liu2016ssd}} 	  &86.67&80.32&48.11&65.35&67.18&72.33&74.45&87.41&80.64&71.22&47.36&63.26&67.49&57.36&58.89& 	69.23	& {\bf{64}} \\ \hline
%------------------------------------

{RICA \cite{li2017rotation}} 	  &86.97&80.93&46.68&67.47&66.19&71.56&74.33&86.43&80.37&71.42&51.76&64.78&71.35&76.84&56.11& 70.21& 24 \\
%---------------------------
{DRN \cite{pan2020dynamic}}  &89.63&82.71&47.25&64.05&76.20&74.33&{{85.76}}&90.53&86.15&84.82&57.77&61.95&69.34&69.72&58.46& 73.25 &9 \\ 

%-------------------------------------
{FMSSD \cite{wang2019fmssd}}   &89.15&83.51&49.23&69.84&69.32&74.57&77.83&{{90.64}}&83.62&75.28&55.37&67.42&75.31&{{80.72}}&60.36&	73.48& 17\\ 
%-----------------------------
{Pelee \cite{wang2018pelee}}   &87.61&73.84&52.93&73.88&72.32&78.15&76.30&{{90.16}}&79.24&76.13&44.89&68.20&72.63&{{78.81}}&79.36&	73.62 & 28\\ 
%------------------------------------
{BBAVectors \cite{yi2021oriented}}  &88.65&84.07&{{52.14}}&69.58&78.24&80.37&\bf{88.03}&90.82&87.16&86.41&{{56.07}}&65.71&67.02&71.94&{{63.97}}& 75.34 &11 \\ 
%-----------------------
{R3Det \cite{yang2021r3det}}  &89.84&83.79&{{48.23}}&66.85&78.71&\bf{83.36}&87.90&\bf{90.86}&85.44&85.46&{{65.74}}&62.75&67.49&78.83&{{72.64}}&76.53  &10 \\ 
%-----------------------
{Scrdet++ \cite{yang2022scrdet++}}  &\bf{90.12}&\bf{85.23}&{{55.61}}&\bf{74.17}&76.48&73.28&86.11&90.53&\bf{87.30}&\bf{87.24}&{\bf{69.73}}&68.81&73.38&72.65&{{67.43}}&\bf{77.20}  &14 \\ 
%-----------------------
{AFD (ours)}  &{{89.81}}&{{77.68}}&\bf{56.17}&{{70.65}}&{\bf{78.94}}&{{81.62}}&84.28&90.35&{{75.23}}&{{76.90}}&51.65&\bf{75.24}&\bf{75.92}&\bf{82.54}&\bf{86.67}& 	{76.91}	&\bf {30} \\
        
        \hline
\end{tabular}
\end{table*}
%----------------------------------
\subsection{Implementation Details}

While the labels on DOTA objects are in a quadrilateral form, those on NWPU are in the common axis-aligned bounding boxes (BBs). In order to have both options, our AFD method provides both oriented and horizontal BBs (HBB, OBB), in which HBB:$\{x_{min},y_{min},x_{max},y_{max}\}$, OBB:$\{x_{center},y_{center},W, H,\theta \}$, while $W$ is width, $H$ is height and $\theta$ is between $[0, 90^{\circ})$ for all objects. In training, a set of rotating rectangles that appropriately overlap with the given quadrilateral labels provide the OBB ground truth. AFD only generates HBB results for the NWPU, due to the datasets' lack of OBB ground truth. AFD, on the other hand, generates both OBB and HBB outputs for the DOTA.

We compare the performance of our model with those of previous KD methods using a variety of object detection strategies \cite{wang2019distilling, zhang2020improve, guo2021distilling, zhang2021learning, yang2022adaptive} in order to show the efficacy of our approach. Our implementation is on the basis of MMDetection \cite{chen2019mmdetection} and on ImageNet the backbone networks are pretrained. The model is implemented in Pytorch and we use four GeForce RTX3090 GPUs for training with the batch size of 16.

The network is trains with 24 epochs using stochastic gradient descent (SGD). The initial learning rate is 0.02 for FR-CNN and 0.01 for the other modules, which are decrease in the $16^{th}$ and $22^{nd}$ epochs by a factor of 10. The momentum and weight decay are set to 0.9 and 1e-4, respectively. The hyper-parameters of losses are set to ($\kappa=5\times10^{-4}$, $\upsilon=2\times10^{-2}$, $\beta=1\times10^{-1}$, and $\mathcal T=1\times10^{-1}$) in the case of one-stage detectors ($\kappa=6\times10^{-5}$, $\upsilon=4\times10^{-3}$, $\beta=1\times10^{-1}$, and $\mathcal T=4\times10^{-1}$) in the case of two-stage detectors.

%-------------------------
%%%%%%%
\begin{figure*}
  \centering
  \includegraphics[width=0.94\textwidth]{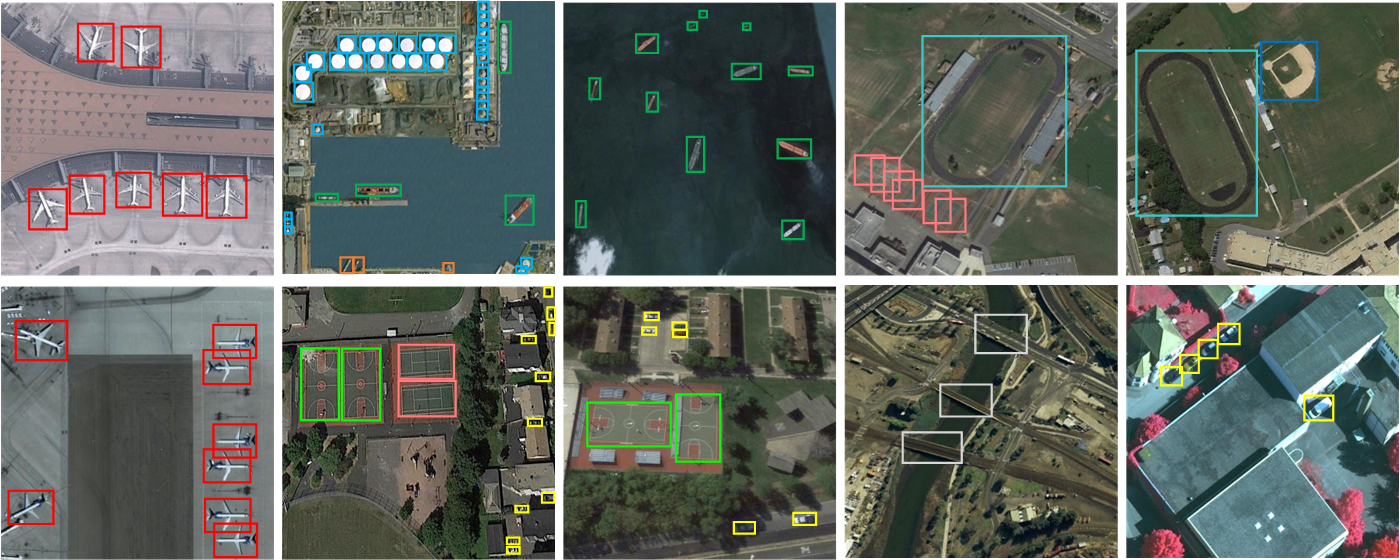}
\caption{Sample detection results of the student using Cascade detector trained with AFD on the NWPU.}
\label{fig:12}
\end{figure*}
%%%%%%%%
%----------------------------------
\begin{table*}
\centering
  \caption {PERFORMANCE COMPARISONS BETWEEN OUR KD DETECTOR WITH CASCADE AND STATE-OF-THE-ART OBJECT DETECTION MODELS ON NWPU DATASET.}
  \label{tab:5}
   \begin{tabular}{m{6em} | p{0.6cm} p{0.6cm} p{0.6cm} p{0.6cm} p{0.6cm} p{0.6cm} p{0.6cm} p{0.7cm} p{0.7cm} p{0.7cm}| p{0.6cm} p{0.5cm}}     \hline \hline
    Methods &Plane&SH&ST&BD&TC&BC&GTF&Harbor&Bridge&Vehicle& mAP & FPS  \\     \hline
%-------------------- 
{EDAI \cite{li2021efficient}}  &77.12&{72.86}&60.34&70.08&59.48&64.82&78.61&67.59&71.94&68.84& 69.17  & 14 \\ 
%--------------------
{RICA \cite{li2017rotation}} &96.43&85.69&89.37&91.48&85.66&78.35&89.30&75.46&69.89&74.96& 83.66 & 31 \\   
%--------------------
{FMSSD \cite{wang2019fmssd}}  &\bf{99.62}&88.71&89.54&97.23&84.65&\bf{95.28}&98.56&73.76&79.43&87.54& 89.43  & 24 \\ 
%--------------------

{Pelee \cite{wang2018pelee}} &99.45&\bf{91.15}&96.21&\bf{97.82}&88.79&90.34&98.30&\bf{86.63}&\bf{86.92}&87.85& 92.34 & 29 \\ 
%--------------------
{AFD (ours)} &{99.51}&90.88&\bf{97.13}&97.36&\bf{89.45}&94.67&\bf{98.95}&85.94&86.34&\bf{88.19}& \bf{92.85} & \bf{35}\\ \hline
\end{tabular}
\end{table*}
%%%%%
%-----------------------------
%-----------------------
\begin{figure}
  \centering
  \includegraphics[width=0.45\textwidth]{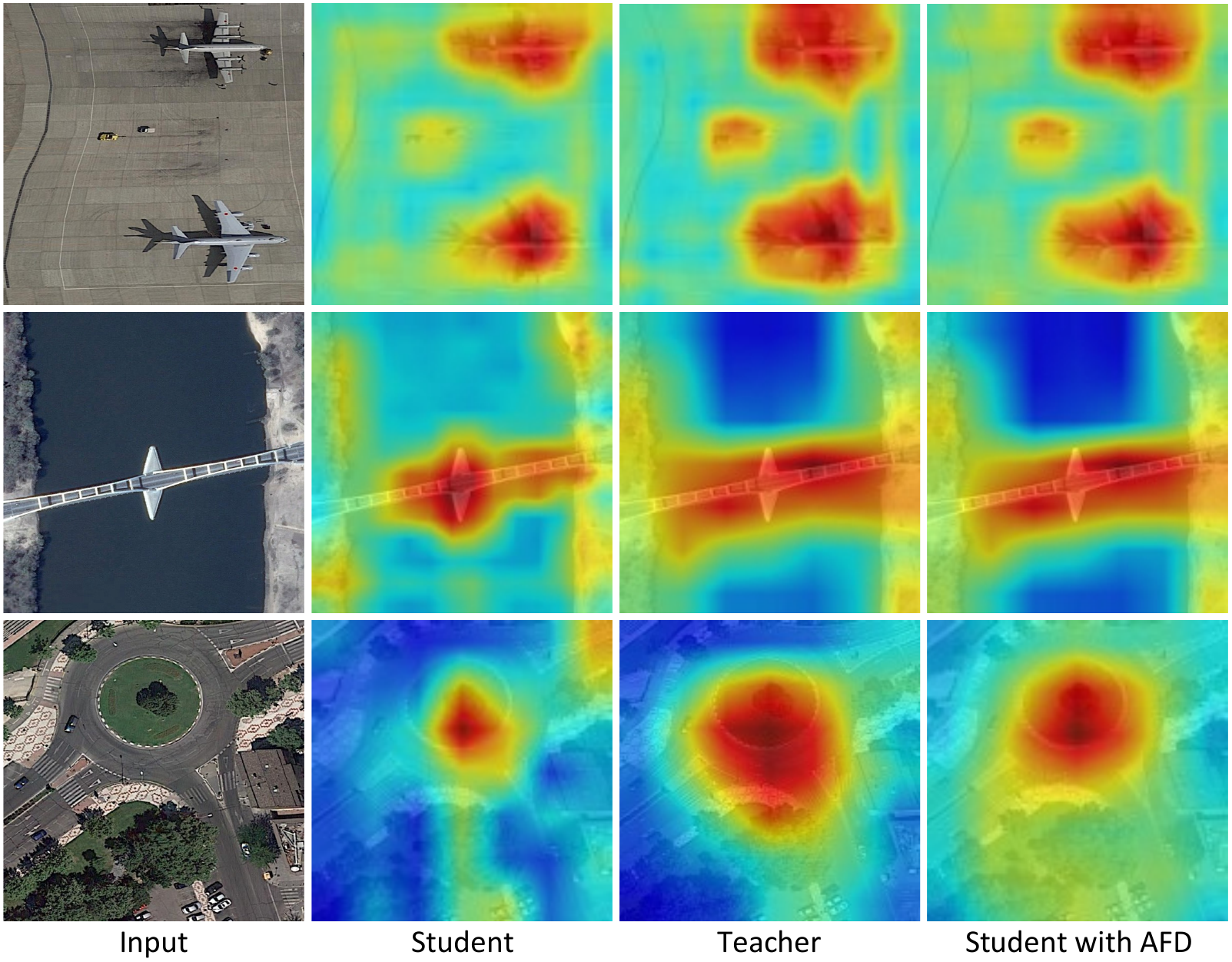}
\caption{Attention maps visualization from different detectors.}
\label{attention}
\end{figure}
%%%%%%%
\subsection{Evaluations using Various Detection Architectures}

We evaluate our AFD's generalization ability on several detection frameworks, including RetinaNet \cite{lin2017focal} (one-stage detector), FR-CNN \cite{ren2016faster} and Cascade \cite{cai2019cascade} (two-stage detectors), and ATSS \cite{zhang2020bridging} and FCOS \cite{tian2019fcos} (anchor-free detectors). We use ResNet18 as student backbones and ResNet101 as teacher backbones for all detectors. As reported in Table \ref{tab.1}, our AFD shows substantial improvements in mAP across various types of detectors. Specifically, when applied to the DOTA dataset, our approach achieves an impressive average mAP enhancement of $5.8$ points, surpassing the performance of standard two-stage detectors. Among the various detectors, our model shows the most significant performance improvement when applied to RetinaNet, enhancing its mAP from an initial $64.47$ to $73.08$. 
Moreover, when AFD used on the NWPU dataset, it achieves remarkable mAP values of $91.85$, $92.67$, and $90.93$ for Cascade, ATSS, and RetinaNet, respectively. The achieved mAP values not only demonstrate competitive performance but also show instances where the student detectors outperform their respective teacher detectors. Notably, in the case of RetinaNet, the incorporation of our AFD yields a marginal yet discernible improvement in student detector performance, attributed to the efficacy of our attention module. These results show that our AFD is adaptable and can be effectively integrated into a wide range of detector architectures, yielding significant performance improvements.

Fig. \ref{attention-2} shows the visualization of attention maps derived from both the student and teacher detectors, as well as from different stages of the student detector using our AFD method. By comparing these attention maps at various training stages, we can observe the student's progressive learning process and its attempt to align with the teacher's guidance.  Fig. \ref{attention-2} demonstrates that the teacher detector shows more accurate focus on the airplanes in the image compared to the student detector, which has undergone only five epochs of training. However, as our AFD progresses, we observe a gradual convergence of attention between the student and teacher. This observation potentially explains why our smaller distillation method even outperforms the teacher model in certain instances (using RetinaNet as detector).

Moreover, Fig. \ref{Fig7} illustrates the progression of classification and localization accuracy during the training process for student detectors using FR-CNN, both with and without the inclusion of our AFD. The data clearly indicates that the incorporation of AFD has a substantial and positive influence on enhancing the performance of the student detector. Notably, the classification accuracy increases from $76.8\%$ without AFD to an impressive $84.3\%$ when AFD is applied.

%-----------------------------
\subsection{Comparative Analysis with Cutting-edge KD Approaches}

On the DOTA dataset we evaluate our model together with recent KD methods using RetinaNet to compare the results with those of other KD approaches. The teacher detector is RetinaNet with ResNet-101, while the student detector is RetinaNet with ResNet-18. 
We conducted a thorough performance evaluation of AFD, comparing it to other state-of-the-art KD models. The results provide clear evidence of AFD's superior performance. As shown in Table \ref{tab.2}, our AFD surpasses all state-of-the-art KD approaches in distillation performance. To be more specific, our model achieved a remarkable improvement, surpassing the dynamic global distillation method \cite{zhang2021learning} by an impressive margin of $1.77$ mAP. This result highlights the superiority of our approach in KD. Moreover, our model achieved an impressive $73.08$ mAP, surpassing the recently developed distillation methods FGD \cite{yang2022focal} and LD \cite{zheng2022localization}, which achieved $71.36$ and $71.43$ mAP, respectively. These results highlight the significant impact of our method's superior extraction of local and global knowledge, resulting in a substantial enhancement in distillation performance.

%------------------------------
\begin{figure}
  \centering
  \includegraphics[width=0.45\textwidth]{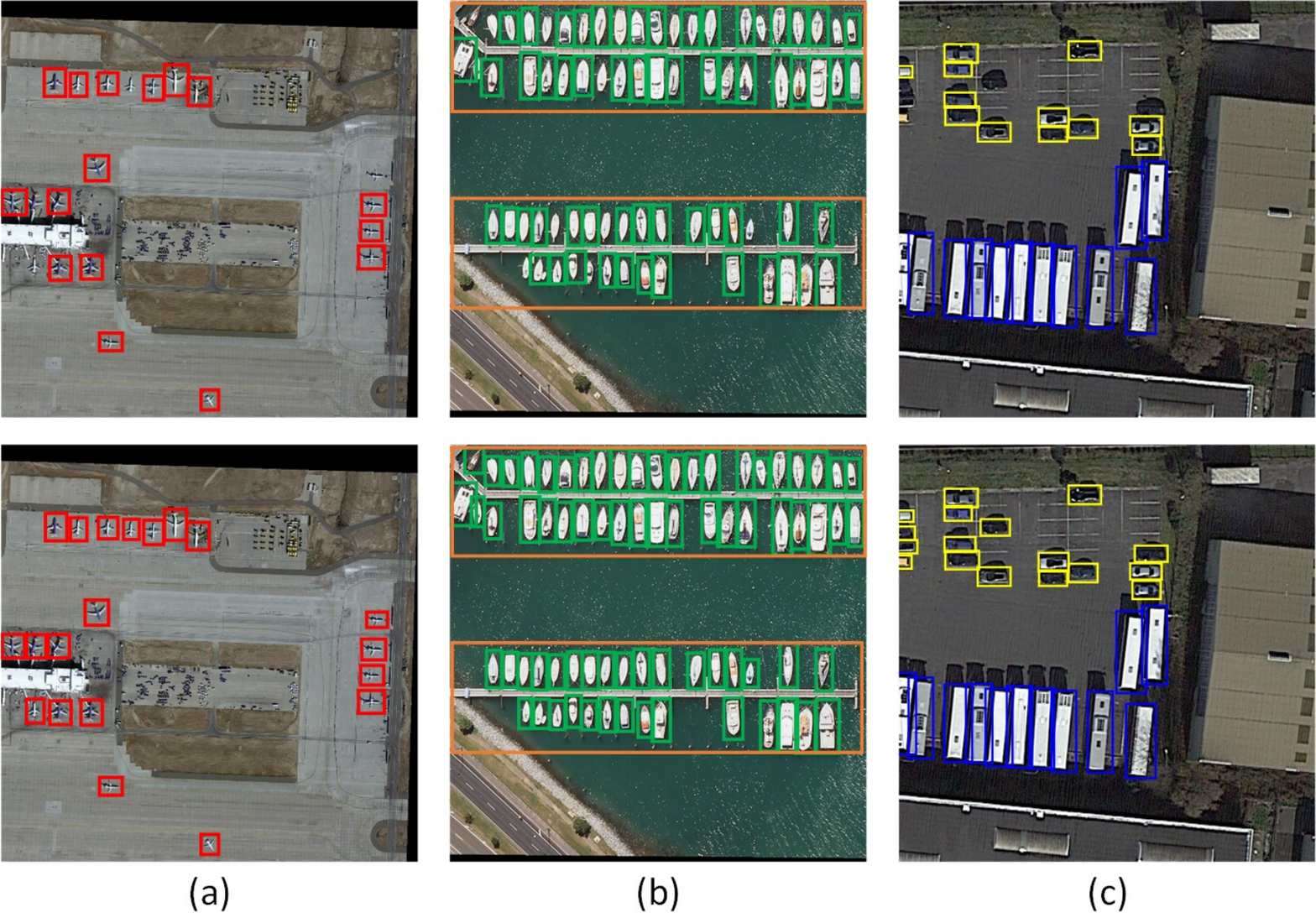}
\caption{The qualitative evaluation of improvement in distillation learning on the DOTA. The top row is the results of student detector without distillation learning and the bottom row is the student that learned by AFD. (a) Plane, (b) ship and harbor, (c) large and small vehicles.}
\label{comp}
\end{figure}
%---------------------
\subsection{Comparative Analysis with CNN-based Detectors}
We evaluate our KD method using the Cascade detector and compare it to other CNN-based object detection models.

In Tables \ref{tab:3} on the DOTA and Table \ref{tab:5} on the NWPU datasets, we respectively evaluate the results of our KD detector compared to the recent object detection models. On the DOTA, FMSSD detects at a rate of $73.48$ mAP and processing $17$ FPS. On the other hand, BBAVectors \cite{yi2021oriented} and R3Det \cite{yang2021r3det} detect at a rate of $75.34$ and $76.53$ mAP while processing at $11$ and $10$ FPS, respectively. Although SCRDET++ \cite{yang2022scrdet++} obtains the highest mAP of $77.20$, its detection speed is just $14$ FPS, which notably lower when compared to the performance of our AFD model.
Empirically, we find our AFD obtains a better detection/speed trade-off compared to other detectors ($76.91$ mAP / $30$ FPS). Some sample detection results of AFD are shown in Fig. \ref{dota}. AFD has the most accurate classification results for the Bridge, Small vehicle, Roundabout, Harbor, Swimming Pool, and Helicopter classes.

As reported in Table \ref{tab:5}, on the NWPU, AFD obtains state-of-the-art results. Some sample detection results of our model are shown in Fig. \ref{fig:12}. Our detector achieves $92.85\%$ mAP with detection speed of $35$ FPS which shows the superiority of AFD compare to other state-of-the-art models for object detection in remote sensing images. AFD detects $6$ and $11$ FPS faster and has a $0.51\%$ and $3.42\%$ higher mAP than Pelee \cite{wang2018pelee} and FMSSD \cite{wang2019fmssd}, respectively. AFD performs best in the Storage tank, Tennis court, Ground track field, and Vehicle classes.

Fig. \ref{attention} illustrates the attention maps produced by the student detector, teacher detector, and student detector with our AFD method using Cascade. Upon observing the attention maps of both the teacher and student detectors, noticeable disparities in pixel distribution become apparent prior to applying the distillation process. However, following training with AFD, the student detector has a similar pixel distribution to the teacher detector, indicating that the student relies on the same regions as the teacher. This shows how AFD improves the performance of the student detector.

The first row in Fig. \ref{comp} shows the baseline visualization results (student detector without KD learning) and the next row represents the results of the student detector learned with our AFD. AFD has a stronger feature extraction ability than the baseline, and the objects are more accurately detected. For example, the detected small vehicles at the bottom of Fig. \ref{comp}(c) show that AFD produces more accurate regression results than the baseline.
%%%%%%%
\begin{figure}
  \centering
  \includegraphics[width=0.49\textwidth]{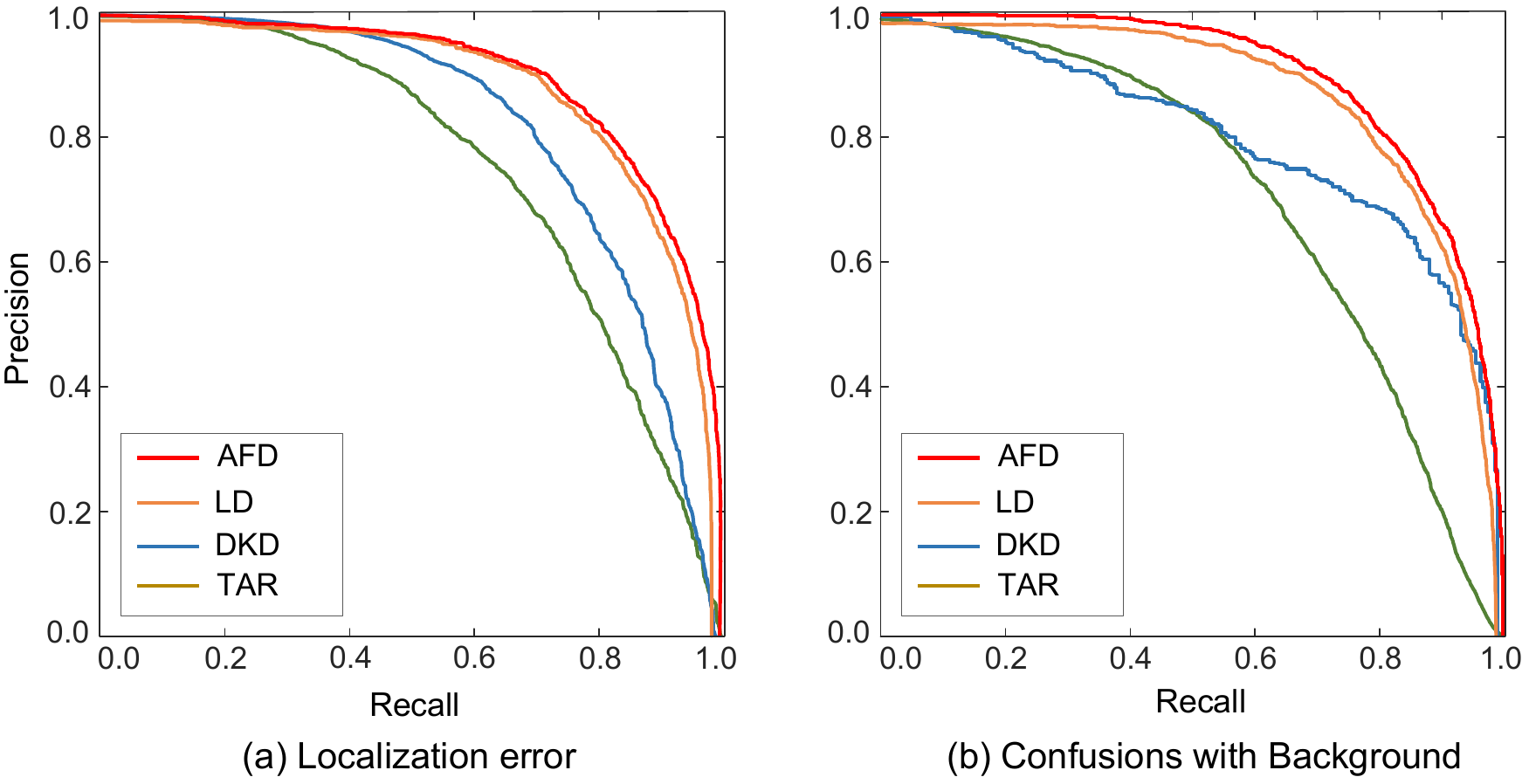}
\caption{Performance analysis of the KD detectors on the DOTA dataset. (a) mean Localization error. (b) Confusions with Background.}
\label{fig:ROC}
\end{figure}
%%%%%
%\vspace{-0.3cm}
To conduct a more comprehensive evaluation of the AFD's performance, in Fig. \ref{fig:ROC} the localization error and confusions with Background curves on the DOTA dataset is shown. As it shows, our model outperforms the other baselines in terms of localisation and classification accuracy. This performance is due to our proposed local and global distillation approach which enables the student detector to understand the relationship between pixels.
The following factors have significantly contributed to this progress.
%%%%%%%%
\begin{itemize} 
\item[1)] The proposed attention feature distillation approach enhances the students' learning of foreground objects while suppressing students' learning of background pixels.
\item[2)] The proposed local and global feature distillation allows the student detector to not only focus on the important pixels and channels of the teacher, but also to recognize the connection between different pixels.
\end{itemize}
%-----------------------------
\subsection{Ablation Study}
In this section, a comprehensive ablation experiments is conducted to assess the importance of the proposed modules of our framework.
%%%%%%%%%%%%%%%
\begin{table}
\centering
  \caption{ABLATION STUDY FOR EACH MODULE'S CONTRIBUTION TO AFD. $\ell_{fd}$, $\ell_{rpn}$, $\ell_{cls-h}$, and $\ell_{loc-h}$ DENOTE DIFFERENT DISTILLATION LOSSES OF OUR METHOD. “LGAM” IS OUR MASK FOR DISTILLATION LOSSES.}
  \label{tab:4}
  \begin{tabular}{p{0.038\textwidth}p{0.036\textwidth}p{0.05\textwidth}p{0.04\textwidth}p{0.04\textwidth}|p{0.03\textwidth}}     \hline \hline
    $\ell_{fd}$ & $\ell_{rpn}$ &LGAM& $\ell_{cls-h}$ &$\ell_{loc-h}$ &mAP  \\     \hline
$\tikzxmark$  & $\hfil \tikzxmark$  &$\hfil \tikzxmark$ & $\hfil \tikzxmark$ & $\hfil \tikzxmark$  & 71.82   \\
$\checkmark$  & $\hfil \tikzxmark$  &$\hfil \tikzxmark$ & $\hfil \tikzxmark$ & $\hfil \tikzxmark$  & 73.41   \\
$\checkmark$  &$\hfil \checkmark$  &$\hfil \tikzxmark$ &$\hfil \tikzxmark$ & $\hfil \tikzxmark$  & 74.63  \\
$\checkmark$  &$\hfil \checkmark$  &$\hfil \checkmark$ &$\hfil \tikzxmark$ &$\hfil \tikzxmark$ & 76.23   \\
$\checkmark$  & $\hfil \checkmark$  &$\hfil \checkmark$ &$\hfil \checkmark$ &$\hfil \tikzxmark$ & 76.56 \\
$\checkmark$  &$\hfil \checkmark$   &$\hfil \checkmark$ &$\hfil \tikzxmark$ &$\hfil \checkmark$ & 76.48 \\
$\checkmark$  & $\hfil \checkmark$  &$\hfil \checkmark$ &$\hfil \checkmark$ &$\hfil \checkmark$ & 76.91  \\ \hline

        \hline
\end{tabular}
\end{table}
%%%%%%%%%%%%%%%
%%%%%%%%%%
\begin{table}
\centering
  \caption {EXPERIMENTS ON HOW PERFORMANCE CHANGES WITH DIFFERENT ATTENTION MASKS AND WITHOUT NORMALIZATION.}
  \label{tab:5-1}
  \begin{tabular}{p{0.042\textwidth}p{0.042\textwidth}p{0.087\textwidth}|p{0.03\textwidth}}     \hline \hline
    Local & Global & Normalization  &mAP  \\     \hline
$\tikzxmark$  & $\hfil \tikzxmark$& $\hfil \tikzxmark$   & 75.54   \\
$\tikzxmark$  & $\hfil \checkmark$& $\hfil \tikzxmark$  & 75.82 \\
$\checkmark$  &$\hfil \tikzxmark$& $\hfil \checkmark$   & 76.65 \\
$\checkmark$  & $\hfil \checkmark$& $\hfil \checkmark$  & 76.91  \\ \hline

        \hline
\end{tabular}
\end{table}
%%%%%%%%%%%%%%%
\subsubsection{AFD Modules}
We compare the AFD detection performance with and without different modules of Eq. (\ref{eq:13}) to evaluate the impact of each of them. Table \ref{tab:4} reports some results of our ablation experiment. On the classification and regression heads, AFD improves mAP by $0.33$ and $0.25$, respectively. When the distillation process is applied jointly to both the classification and regression heads, our model shows an improvement of $0.68$ mAP. These results provide clear evidence that each component of our total distillation loss significantly contributes to the overall enhancement.
%%%%%%%%%%
%%%%%%%
\begin{figure}
  \centering
  \includegraphics[width=0.49\textwidth]{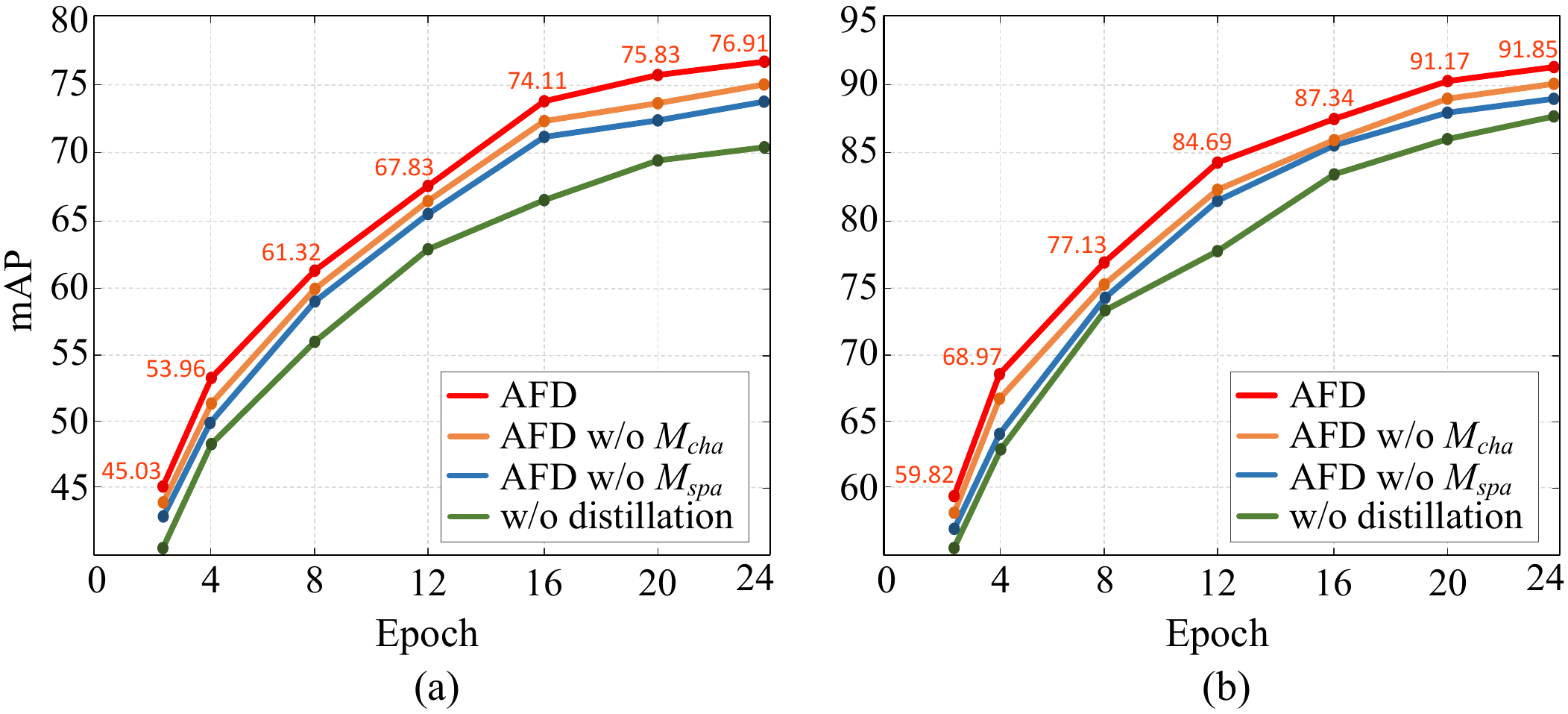}
\caption{The mAP during training. (a) Performance evaluation on DOTA. (b) Performance evaluation on NWPU. Channel and spatial masks can improve mAP in training stages.}
\label{fig6}
\end{figure}
%%%%%
%%%%%%%%%%
\begin{table}
\centering
  \caption{EFFECT OF SPATIAL AND CHANNEL ATTENTION.}
  \label{tab:7}
  \begin{tabular}{p{0.062\textwidth}p{0.07\textwidth}|p{0.03\textwidth}}     \hline \hline
    Spatial attention & Channel attention   &mAP  \\     \hline
$\tikzxmark$  & $\hfil \tikzxmark$   & 73.44   \\
$\tikzxmark$  & $\hfil \checkmark$  & 76.73 \\
$\checkmark$  &$\hfil \tikzxmark$   & 76.08 \\
$\checkmark$  & $\hfil \checkmark$  & 76.91  \\ \hline

        \hline
\end{tabular}
\end{table}
%%%%%%%%%%%%%%%
%%%%%%%%%%
\begin{table}[t]
\centering
  \caption{THE RESULTS OF VARIOUS LOSS FUNCTIONS UTILIZED IN $\ell_{loc-h}$. }
  \label{tab:8}
  \begin{tabular}{p{0.047\textwidth}|p{0.047\textwidth}p{0.085\textwidth}p{0.047\textwidth}p{0.047\textwidth}}     \hline \hline
    Loss & $\ell_1$ & $smooth-\ell_1$  &$\ell_{MSE}$& $\ell_{IoU}$  \\     \hline
$mAP$  & 76.84& $\hfil$ 76.69 & 76.72&76.91  \\ \hline

        \hline
\end{tabular}
\end{table}
%%%%%%%%%%%%%%%
\subsubsection{Distillation Effect of Local and Global Attention Masks}
Since the global attention mask is not distributed equally, KD only with the global feature attention module can extract information from large objects. Alternatively, we believe that small objects can be retrieved using local feature attention. We conduct different distillations using local and global attention masks to determine which is more effective. Based on the data in Table \ref{tab:5-1}, it is clear that local attention leads to a higher mAP than global attention. Despite that, the global attention results are inferior to the local ones, however, detection performance further improves when both local and global attention are used together. Following these observations, AFD's local attention mask should work with the global attention mask to boost performance further. 
Furthermore, by using normalization process, a large uniform value of loss weight can be obtained between the teacher and student detectors to maintain the balance of the detection and distillation losses. This balance is important to ensure that both the detection loss and the distillation loss have appropriate contributions to the overall learning process. Without normalization, one of the losses (detection or distillation) might dominate the training process due to large magnitude differences. This can lead to an imbalance in learning objectives and hinder the effectiveness of KD. Normalization mitigates this issue by ensuring that both losses have comparable magnitudes and influence the training process in a more balanced manner.  
%%%%%%%%%%%%%

\subsubsection{Analysis of Different Attentions and Distillation Losses}
As shown in Table \ref{tab:7}, spatial and channel attentions boost mAP by $3.29$ and $2.64$, respectively. On the other hand, combining the two attentions result in $3.47$ mAP improvements. The findings indicate the valuable contributions of both channel attention and spatial attention, highlighting their effective combination as a means to enhance overall performance.
Indeed, the teacher detector effectively guides the student detector's focus towards important components by using a spatial and channel attention mask. We analyze the impact of different masks in Fig. \ref{fig6}. Each attention mask increases our model efficiency, particularly the spatial attention mask. However, the best result is obtained by combining both the masks.
Additionally, we evaluate the effects of the various loss functions of Eq. (\ref{eq:12}). In order to evaluate the regression loss in Eq. (\ref{eq:12}), we analyze the result of several losses such as $\ell_{IoU}$, $\ell_{MSE}$, $\ell_1$, and $smooth-\ell_1$. The $\ell_{IoU}$ has the best performance compared to the others, as reported in Table \ref{tab:8}.

%------------------
\section{Conclusion}
\label{sec:5}
This paper introduced AFD, a new mask-based KD approach for target detection in remote sensing images that efficiently uses local and global attention methods to obtain local features and background information. To extract local features, we split the feature maps of input image into patches and apply attention methods. Our method enhances distillation performance by extracting both fine-grained features and more important background information from a range of objects. We showed that AFD outperforms other KD techniques when combined with the different detection systems. The detection results demonstrate that AFD surpasses state-of-the-art models performance and can be adopted in various detectors such as single-stage, two-stage, and even anchor-free. Moreover, we conducted an ablation experiment and analysis, demonstrating the importance of distilling local information from multiple regions for object detection. We believe our work represents a turning point for traditional KD approaches that only rely on global information, into more effective model that incorporate both local and global information.

\footnotesize{
\bibliographystyle{IEEEtran}
\bibliography{IEEEabrv,IEEEexample}
}

%\begin{thebibliography}{1}
%\bibitem{IEEEhowto:kopka}
%[49] Denton, Emily L and Chintala, Soumith and Fergus, Rob and others Deep Generative Image Models using a ￼Laplacian Pyramid of Adversarial Networks}
%\end{thebibliography}

\end{document}